%% file: semeval2018.tex
\documentclass[11pt,a4paper]{article}
\usepackage[hyperref]{naaclhlt2018}
\usepackage{times}
\usepackage{latexsym}
\usepackage{booktabs}
\usepackage{array}
\usepackage{pgf}
\usepackage{url}
\usepackage{amsmath}

\aclfinalcopy 

\title{GU IRLAB at SemEval-2018 Task 7: \\Tree-LSTMs for Scientific Relation Classification}

\author{Sean MacAvaney, Luca Soldaini, Arman Cohan, and Nazli Goharian \\
  Information Retrieval Lab \\
  Department of Computer Science \\
  Georgetown University \\
  {\tt \{firstname\}@ir.cs.georgetown.edu}}

\date{}

\begin{document}
\maketitle
\begin{abstract}
SemEval 2018 Task 7 focuses on relation extraction and classification in scientific literature. In this work, we present our tree-based LSTM network for this shared task. Our approach placed 9th (of 28) for subtask 1.1 (relation classification), and 5th (of 20) for subtask 1.2 (relation classification with noisy entities). We also provide an ablation study of features included as input to the network.
\end{abstract}

\section{Introduction}

Information Extraction (IE) has applications in a variety of domains, including in scientific literature. Extracted entities and relations from scientific articles could be used for a variety of tasks, including abstractive summarization, identification of articles that make similar or contrastive claims, and filtering based on article topics. While ontological resources can be leveraged for entity extraction~\citep{Gbor2016SemanticAO}, relation extraction and classification still remains a challenging task. Relations are particularly valuable because (unlike simple entity occurrences) relations between entities capture lexical semantics. SemEval 2018 Task 7 (Semantic Relation Extraction and Classification in Scientific Papers) encourages research in relation extraction in scientific literature by providing common training and evaluation datasets~\cite{SemEval2018Task7}. In this work, we describe our approach using a tree-structured recursive neural network, and provide an analysis of its performance.

There has been considerable previous work with scientific literature due to its availability and interest to the research community. A previous shared task (SemEval 2017 Task 10) investigated the extraction of both keyphrases (entities) and relations in scientific literature~\cite{Augenstein2017SemEval2T}. However, the relation set for this shared task was limited to just synonym and hypernym relationships. The top three approaches used for relation-only extraction included convolutional neural networks~\cite{lee-dernoncourt-szolovits:2017:SemEval}, bi-directional recurrent neural networks with Long Short-Term Memory (LSTM, \citeauthor{hochreiter1997long}, \citeyear{hochreiter1997long}) cells~\cite{Ammar2017TheAS}, and conditional random fields~\cite{Lee2017TheNS}.

There are several challenges related to scientific relation extraction. One is the extraction of the entities themselves. \citet{Luan2017ScientificIE} produce the best published results on the 2017 \mbox{ScienceIE} shared task for entity extraction using a semi-supervised approach with a bidirectional LSTM and a CRF tagger. \citet{Zheng2014EntityLF} provide an unsupervised technique for entity linking scientific entities in the biomedical domain to an ontology.

\textbf{Contribution.} Our approach employs a tree-based LSTM network using a variety of syntactic features to perform relation label classification. We rank 9th (of 28) when manual entities are used for training, and 5th (of 20) when noisy entities are used for training. Furthermore, we provide an ablation analysis of the features used by our model. Code for our model and experiments is available.\footnote{\url{https://github.com/Georgetown-IR-Lab/semeval2018-task7}}

\section{Methodology}

Syntactic information between entities plays an important role in relation extraction and classification \citep{mintz2009distant,macavaney2017guir}.
Similarly, sequential neural models, such as LSTM, have shown promising results on scientific literature~\citep{Ammar2017TheAS}.
Therefore, in our approach, we leverage both syntactic structures and neural sequential models by employing a tree-based long-short term memory cell (tree-LSTM). 
Tree-LSTMs, originally introduced by~\citet{tai2015improved}, have been successfully used to capture relation information in other domains~\citep{xu2015classifying,miwa2016end}. 
On a high level, tree-LSTMs operate very similarly to sequential models; however, rather than processing tokens sequentially, they follow syntactic dependencies; once the model reaches the root of the tree, the output is used to compute a prediction, usually through a dense layer.  
We use the child-sum variant of tree-LSTM \cite{tai2015improved}.

\begin{figure}
\centering
\includegraphics[scale=2.8]{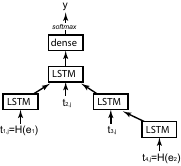}
\caption{Our tree LSTM network.}\label{fig:network}
\end{figure}

Formally, let $S_j = \{t_{1,j},\ldots, t_{n,j}\}$ be a sentence of length $n$, $e_1 = \{t_{i},\ldots, t_{k}\}$ and $e_2 = \{t_{p},\ldots, t_{q}\}$ two entities whose relationship we intend to classify; 
let $\text{H}(e_1)$, $\text{H}(e_2)$ be the root of the syntactic subtree spanning over entities $e_1$ and $e_2$. 
Finally, let $\text{T}(e_1, e_2)$ be the syntactic sub-tree spanning from $\text{H}(e_1)$ to $\text{H}(e_2)$. 
For the first example in Table~\ref{tab:intro-examples}, $e_1=\{\textit{`Oral', `communication'}\}$ , $e_2=\{\textit{`indices'}\}$, $\text{H}(e_1)=\{\textit{`communication'}\}$, $\text{T}(e_1, e_2)=\{\textit{`communication', `offer', `indices'}\}$.
The proposed model uses word embeddings of terms in $\text{T}(e_1, e_2)$ as inputs; the output of the tree-LSTM cell on the root of the syntactic tree is used to predict one of the six relation types ($y$) using a softmax layer. A diagram of our tree LSTM network is shown in Figure~\ref{fig:network}.

In order to overcome the limitation imposed by the small amount of training data available for this task, we modify the general architecture proposed in \cite{miwa2016end} in two crucial ways.
First, rather than using the representation of entities as input, we only consider the syntactic head of each entity. 
This approach improves the generalizability of the model, as it prevents overfitting on very specific entities in the corpus. 
For example, by reducing \textit{`Bag-of-words methods'} to \textit{`methods'} and \textit{`segment order-sensitive models'} to \textit{`models'}, the model is able to recognize the \textsc{compare} relation between these two entities (see Table~\ref{tab:intro-examples}).
Second, we experimented with augmenting each term representation with the following features:
\begin{itemize}
  \item Dependency labels (\textsc{DEP}): we append to each term embedding the label representing the dependency between the term and its parent.
  \item PoS tags (\textsc{POS}): the part-of-speech tag for each term is append to its embedding. 
  \item Entity length (\textsc{EntLen}): we concatenate the number of tokens in $e_1$ and $e_2$ to embeddings representation of heads $\text{H}(e_1)$ to $\text{H}(e_2)$. For terms that are not entity heads, the entity length feature is replaced by `0'.
  \item Height: the height of each term in the syntactic subtree connecting two entities. 
\end{itemize}

\section{Experimental Setup}

\begin{table}
\small
\renewcommand{\arraystretch}{1.4}
\begin{tabular}{>{\raggedright\arraybackslash}p{2.4cm}>{\raggedright\arraybackslash}p{4.4cm}}
\toprule
Relation (abbr.) & Example \\
\midrule
\textsc{usage} (U) & \underline{Oral communication} may offer additional \underline{indices}... \\
\textsc{model-feature} (M-F) & We look at the \underline{intelligibility} of \underline{MT output}... \\
\textsc{part\_whole} (P-W) & As the \underline{operational semantics} of \underline{natural language applications} improve... \\
\textsc{compare} (C) & \underline{Bag-of-words methods} are shown to be equivalent to \underline{segment order-sensitive methods} in terms of... \\
\textsc{result} (R) & We find that \underline{interpolation methods} improve the \underline{performance}... \\
\textsc{topic} (T) & A \underline{formal analysis} for a large class of words called \underline{alternative markers}... \\
\toprule
\end{tabular}
\caption{Example relations for each type. Entities are underlined, and all relations are from the first entity to the second entity (non-reversed).}\label{tab:intro-examples}
\end{table}

\begin{table}
\centering
\begin{tabular}{lrrrrrr}
\toprule
Dataset & U & M-F & P-W & C & R & T \\
\midrule
\multicolumn{2}{l}{\textbf{Subtask 1.1}} \\
Train & 409&289&215&86&57&15 \\
Valid. & 74&37&19&9&15&3 \\
Test & 175&66&70&21&20&3 \\
\midrule
\multicolumn{2}{l}{\textbf{Subtask 1.2}} \\
Train & 363&124&162&29&94&207 \\
Valid. & 107&51&34&12&29&36 \\
Test & 123&75&56&3&29&69 \\
\bottomrule
\end{tabular}
\caption{Frequency of relation labels in train, validation, and test sets. See Table~\ref{tab:intro-examples} for relation label abbreviations. Subtask 1.1 uses manual entity labels, and subtask 1.2 uses automatic entity labels (which may be noisy).}\label{tab:datasets}
\end{table}

SemEval 2018 Task 7 focuses on relation extraction, assuming a gold set of entities. This allows participants to focus on specific issues related to relation extraction with a rich set of semantic relations. These include relations for \textsc{usage}, \textsc{model-feature}, \textsc{part\_whole}, \textsc{compare}, \textsc{result}, and \textsc{topic}. Examples of each type of relation are given in Table~\ref{tab:intro-examples}.

The shared task evaluates three separate subtasks (1.1, 1.2, and 2). We tuned and submitted our system for subtasks 1.1 and 1.2. In both of these subtasks, participants are given scientific abstracts with entities and candidate relation pairs, and are asked to determine the relation label of each pair. For subtask 1.1, both the entities and relations are manually annotated. For subtask 1.2, the entities are automatically generated using the procedure described in~\citet{Gbor2016SemanticAO}. This procedure introduces noise, but represents a more realistic evaluation environment than subtask 1.1. In both cases, relations and gold labels are produced by human annotators. All abstracts are from the ACL Anthology Reference Corpus~\cite{Bird2008TheAA}. We randomly select 50 texts from the training datasets for validation of our system. We provide a summary of the datasets for training, validation, and testing in Table~\ref{tab:datasets}. Notice how the proportions of each relation label vary considerably among the datasets.

We experiment with two sets of word embeddings: Wiki News and arXiv. The Wiki News embeddings benefit from the large amount of general language, and the arXiv embeddings capture specialized domain language. The Wiki News embeddings are pretrained using fastText with a dimension of 300~\cite{mikolov2018advances}. The arXiv embeddings are trained on a corpus of text from the cs section of arXiv.org\footnote{\url{https://github.com/acohan/long-summarization}} using a window of 8 (to capture adequate term context) and a dimension of 100~\cite{cohan2018}. A third variation of the embeddings simply concatenates the Wiki News and arXiv embeddings, yielding a dimension of 400; for words that appear in only one of the two embedding sources, the available embeddings are concatenated with a vector of appropriate size sampled from $\mathcal{N}(0, 10^{-8})$.

For our official SemEval submission, we train our model using the concatenated embeddings and one-hot encoded dependency label features. We use a hidden layer of 200 nodes, a 0.2 dropout rate, and a training batch size of 16.
Syntactic trees were extracted using SpaCy\footnote{\url{https://spacy.io/}}, and the neural model was implemented using MxNet\footnote{\url{https://mxnet.incubator.apache.org/}}.

The official evaluation metric is the macro-averaged F1 score of all relation labels. For additional analysis, we use the macro precision and recall, and the F1 score for each relation label.

\section{Results and Discussion}

\begin{table}
\centering
\begin{tabular}{lrr}
\toprule
System & F1 & Rank \\
\midrule
\textbf{Subtask 1.1} (28 teams) \\
Our submission & 60.9 & 9 \\
Median team & 45.5 \\
Mean team & 37.1 \\
\midrule
\textbf{Subtask 1.2} (20 teams) \\
Our submission & 78.9 & 5 \\
Median team & 70.3 \\
Mean team & 54.0 \\
\bottomrule
\end{tabular}
\caption{Performance result comparison to other task participants for subtasks 1.1 and 1.2.}\label{tab:results-compare}
\end{table}

\begin{table*}[h]
\centering
\begin{tabular}{lrrrrrrrrr}
\toprule
&\multicolumn{3}{c}{Overall} & \multicolumn{6}{c}{F1 by label} \\ \cmidrule(lr){2-4}\cmidrule(lr){5-10}
Features&P&R&F1&U&M-F&P-W&C&R&T\\
\midrule
\textbf{Subtask 1.1} \\
(no features)&56.9&64.1&59.5&\bf81.4&51.5&59.9&57.8&61.9&44.4\\
DEP&53.5&54.1&53.6&79.1&55.5&58.2&63.8&\bf64.9&0.0\\
DEP + POS&\bf60.1&59.1&59.5&79.9&57.1&58.5&\bf68.3&60.0&33.3\\
DEP + POS + EntLen&59.4&\bf64.1&\bf60.9&80.0&\bf59.0&56.9&58.3&61.1&\bf50.0\\
DEP + POS + EntLen + Height&52.1&53.3&52.4&79.2&57.4&\bf62.2&56.0&59.5&0.0\\
\midrule
\textbf{Subtask 1.2} \\
(no features)&74.2&78.9&75.4&80.0&65.6&72.6&57.1&80.0&\bf97.1\\
DEP&76.4&78.5&76.4&79.2&67.2&73.0&\bf66.7&79.4&93.1\\
DEP + POS&75.5&\bf80.3&77.3&\bf82.0&\bf73.9&\bf73.6&57.1&80.0&\bf97.1\\
DEP + POS + EntLen&\bf78.2&79.7&\bf78.0&81.9&69.3&70.5&\bf66.7&\bf82.5&\bf97.1\\
DEP + POS + EntLen + Height&73.0&78.7&74.8&79.5&70.7&70.3&57.1&74.3&\bf97.1\\
\bottomrule
\end{tabular}
\caption{Feature ablation results for subtasks 1.1 and 1.2. DEP are dependency labels, POS are part of speech labels, EntLen is is the length of the input entities, and Height is the height of the entities in the dependency tree. In both subtasks 1.1 and 1.2, the combination of dependency labels, parts of speech, and entity lengths yield the best performance in terms of overall F1 score.}\label{tab:results-more}\label{tab:results-features}
\end{table*}
\renewcommand{\arraystretch}{0.9}
\begin{table}
\centering
\begin{tabular}{lrrr}
\toprule
Embeddings & P & R & F1 \\
\midrule
\textbf{Subtask 1.1} \\
Wiki News &59.2&57.3&57.6 \\
arXiv & 58.5&55.1&56.4\\
Wiki News + arXiv &\bf59.4&\bf64.1&\bf60.9 \\
\midrule
\textbf{Subtask 1.2} \\
Wiki News & 73.1&76.2&72.7 \\
arXiv & 65.4&67.4&65.9 \\
Wiki News + arXiv &\bf78.2&\bf79.7&\bf78.0 \\
\bottomrule
\end{tabular}
\caption{Performance comparison for subtasks 1.1 and 1.2 when using Wiki News and arXiv embeddings. The concatenated embeddings outperform the individual methods.}\label{tab:results-embed}
\vspace{-1em}
\end{table}

\begin{figure}
\centering
\vspace{-0.05in}
\scalebox{0.46}{\input{confmat}}
\caption{Confusion matrix for subtask 1.1.}\label{fig:confmat}
\end{figure}

In Table~\ref{tab:results-compare}, we provide our official SemEval results in the context of other task participants. In both subtasks, we ranked above both the median and mean team scores, treating the top-ranking approach for each team as the team's score. For Subtask 1.1, we ranked 9 out of 28, and for Subtask 1.2, we ranked 5 out of 20. This indicates that our approach is generally more tolerant to the noisy entities given in Subtask 1.2 than most other approaches. Figure~\ref{fig:confmat} is a confusion matrix for the official submission for subtask 1.1. The three most frequent labels in the training data (\textsc{usage}, \textsc{model-feature}, and \textsc{part\_whole}) are also the most frequently confused relation labels. This behavior can be partially attributed to the class imbalance.

In Table~\ref{tab:results-features}, we examine the effects of various feature combinations on the model. Specifically, we check the macro averaged precision, recall, and F1 scores for both subtask 1.1 and 1.2 with various sets of features on the test set. Of the combinations we investigated, including the dependency labels, part of speech tags, and the token length of entities yielded the best results in terms of overall F1 score for both subtasks. The results by individual relation label are more mixed, with the overall best combination simply yielding better performance on average, not on each label individually. Interestingly, the entity height feature reduces performance, perhaps indicating that it is easy to overfit the model using this feature.

Table~\ref{tab:results-embed} examines the effect of the choice of word embeddings on performance. In both subtasks, concatenating the Wiki News and arXiv embeddings yields better performance than using a single type of embedding. This suggests that the two types of embeddings are useful in different cases; perhaps Wiki News better captures the general language linking the entities, whereas the arXiv embeddings capture the specialized language of the entities themselves.

\section{Conclusion}

In this work, we investigated the use of a tree LSTM-based approach for relation classification in scientific literature. Our results at SemEval 2018 were encouraging, placing 9th (of 28) at subtask 1.1 (relation classification with manually-annotated entities), and 5th (of 20) at subtask 1.2 (relation classification using automatically-generated entities). Furthermore, we conducted an analysis of our system by varying the system parameters and features.

\bibliography{semeval2018}
\bibliographystyle{acl_natbib}

\end{document}

%% file: confmat.tex
\begingroup%
\makeatletter%
\begin{pgfpicture}%
\pgfpathrectangle{\pgfpointorigin}{\pgfqpoint{4.000000in}{4.000000in}}%
\pgfusepath{use as bounding box, clip}%
\begin{pgfscope}%
\pgfsetbuttcap%
\pgfsetmiterjoin%
\definecolor{currentfill}{rgb}{1.000000,1.000000,1.000000}%
\pgfsetfillcolor{currentfill}%
\pgfsetlinewidth{0.000000pt}%
\definecolor{currentstroke}{rgb}{1.000000,1.000000,1.000000}%
\pgfsetstrokecolor{currentstroke}%
\pgfsetdash{}{0pt}%
\pgfpathmoveto{\pgfqpoint{0.000000in}{0.000000in}}%
\pgfpathlineto{\pgfqpoint{4.000000in}{0.000000in}}%
\pgfpathlineto{\pgfqpoint{4.000000in}{4.000000in}}%
\pgfpathlineto{\pgfqpoint{0.000000in}{4.000000in}}%
\pgfpathclose%
\pgfusepath{fill}%
\end{pgfscope}%
\begin{pgfscope}%
\pgfsetbuttcap%
\pgfsetmiterjoin%
\definecolor{currentfill}{rgb}{1.000000,1.000000,1.000000}%
\pgfsetfillcolor{currentfill}%
\pgfsetlinewidth{0.000000pt}%
\definecolor{currentstroke}{rgb}{0.000000,0.000000,0.000000}%
\pgfsetstrokecolor{currentstroke}%
\pgfsetstrokeopacity{0.000000}%
\pgfsetdash{}{0pt}%
\pgfpathmoveto{\pgfqpoint{0.560556in}{0.454306in}}%
\pgfpathlineto{\pgfqpoint{3.820000in}{0.454306in}}%
\pgfpathlineto{\pgfqpoint{3.820000in}{3.713750in}}%
\pgfpathlineto{\pgfqpoint{0.560556in}{3.713750in}}%
\pgfpathclose%
\pgfusepath{fill}%
\end{pgfscope}%
\begin{pgfscope}%
\pgfpathrectangle{\pgfqpoint{0.560556in}{0.454306in}}{\pgfqpoint{3.259444in}{3.259444in}} %
\pgfusepath{clip}%
\pgftext[at=\pgfqpoint{0.560556in}{0.454306in},left,bottom]{\pgfimage[interpolate=true,width=3.270000in,height=3.270000in]{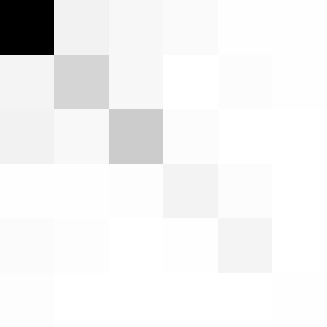}}%
\end{pgfscope}%
\begin{pgfscope}%
\pgfsetrectcap%
\pgfsetmiterjoin%
\pgfsetlinewidth{1.003750pt}%
\definecolor{currentstroke}{rgb}{0.000000,0.000000,0.000000}%
\pgfsetstrokecolor{currentstroke}%
\pgfsetdash{}{0pt}%
\pgfpathmoveto{\pgfqpoint{0.560556in}{0.454306in}}%
\pgfpathlineto{\pgfqpoint{0.560556in}{3.713750in}}%
\pgfusepath{stroke}%
\end{pgfscope}%
\begin{pgfscope}%
\pgfsetrectcap%
\pgfsetmiterjoin%
\pgfsetlinewidth{1.003750pt}%
\definecolor{currentstroke}{rgb}{0.000000,0.000000,0.000000}%
\pgfsetstrokecolor{currentstroke}%
\pgfsetdash{}{0pt}%
\pgfpathmoveto{\pgfqpoint{0.560556in}{3.713750in}}%
\pgfpathlineto{\pgfqpoint{3.820000in}{3.713750in}}%
\pgfusepath{stroke}%
\end{pgfscope}%
\begin{pgfscope}%
\pgfsetrectcap%
\pgfsetmiterjoin%
\pgfsetlinewidth{1.003750pt}%
\definecolor{currentstroke}{rgb}{0.000000,0.000000,0.000000}%
\pgfsetstrokecolor{currentstroke}%
\pgfsetdash{}{0pt}%
\pgfpathmoveto{\pgfqpoint{0.560556in}{0.454306in}}%
\pgfpathlineto{\pgfqpoint{3.820000in}{0.454306in}}%
\pgfusepath{stroke}%
\end{pgfscope}%
\begin{pgfscope}%
\pgfsetrectcap%
\pgfsetmiterjoin%
\pgfsetlinewidth{1.003750pt}%
\definecolor{currentstroke}{rgb}{0.000000,0.000000,0.000000}%
\pgfsetstrokecolor{currentstroke}%
\pgfsetdash{}{0pt}%
\pgfpathmoveto{\pgfqpoint{3.820000in}{0.454306in}}%
\pgfpathlineto{\pgfqpoint{3.820000in}{3.713750in}}%
\pgfusepath{stroke}%
\end{pgfscope}%
\begin{pgfscope}%
\pgfsetbuttcap%
\pgfsetroundjoin%
\definecolor{currentfill}{rgb}{0.000000,0.000000,0.000000}%
\pgfsetfillcolor{currentfill}%
\pgfsetlinewidth{0.501875pt}%
\definecolor{currentstroke}{rgb}{0.000000,0.000000,0.000000}%
\pgfsetstrokecolor{currentstroke}%
\pgfsetdash{}{0pt}%
\pgfsys@defobject{currentmarker}{\pgfqpoint{0.000000in}{0.000000in}}{\pgfqpoint{0.000000in}{0.055556in}}{%
\pgfpathmoveto{\pgfqpoint{0.000000in}{0.000000in}}%
\pgfpathlineto{\pgfqpoint{0.000000in}{0.055556in}}%
\pgfusepath{stroke,fill}%
}%
\begin{pgfscope}%
\pgfsys@transformshift{0.832176in}{0.454306in}%
\pgfsys@useobject{currentmarker}{}%
\end{pgfscope}%
\end{pgfscope}%
\begin{pgfscope}%
\pgfsetbuttcap%
\pgfsetroundjoin%
\definecolor{currentfill}{rgb}{0.000000,0.000000,0.000000}%
\pgfsetfillcolor{currentfill}%
\pgfsetlinewidth{0.501875pt}%
\definecolor{currentstroke}{rgb}{0.000000,0.000000,0.000000}%
\pgfsetstrokecolor{currentstroke}%
\pgfsetdash{}{0pt}%
\pgfsys@defobject{currentmarker}{\pgfqpoint{0.000000in}{-0.055556in}}{\pgfqpoint{0.000000in}{0.000000in}}{%
\pgfpathmoveto{\pgfqpoint{0.000000in}{0.000000in}}%
\pgfpathlineto{\pgfqpoint{0.000000in}{-0.055556in}}%
\pgfusepath{stroke,fill}%
}%
\begin{pgfscope}%
\pgfsys@transformshift{0.832176in}{3.713750in}%
\pgfsys@useobject{currentmarker}{}%
\end{pgfscope}%
\end{pgfscope}%
\begin{pgfscope}%
\pgftext[x=0.832176in,y=0.398750in,,top]{\sffamily\fontsize{12.000000}{14.400000}\selectfont U}%
\end{pgfscope}%
\begin{pgfscope}%
\pgfsetbuttcap%
\pgfsetroundjoin%
\definecolor{currentfill}{rgb}{0.000000,0.000000,0.000000}%
\pgfsetfillcolor{currentfill}%
\pgfsetlinewidth{0.501875pt}%
\definecolor{currentstroke}{rgb}{0.000000,0.000000,0.000000}%
\pgfsetstrokecolor{currentstroke}%
\pgfsetdash{}{0pt}%
\pgfsys@defobject{currentmarker}{\pgfqpoint{0.000000in}{0.000000in}}{\pgfqpoint{0.000000in}{0.055556in}}{%
\pgfpathmoveto{\pgfqpoint{0.000000in}{0.000000in}}%
\pgfpathlineto{\pgfqpoint{0.000000in}{0.055556in}}%
\pgfusepath{stroke,fill}%
}%
\begin{pgfscope}%
\pgfsys@transformshift{1.375417in}{0.454306in}%
\pgfsys@useobject{currentmarker}{}%
\end{pgfscope}%
\end{pgfscope}%
\begin{pgfscope}%
\pgfsetbuttcap%
\pgfsetroundjoin%
\definecolor{currentfill}{rgb}{0.000000,0.000000,0.000000}%
\pgfsetfillcolor{currentfill}%
\pgfsetlinewidth{0.501875pt}%
\definecolor{currentstroke}{rgb}{0.000000,0.000000,0.000000}%
\pgfsetstrokecolor{currentstroke}%
\pgfsetdash{}{0pt}%
\pgfsys@defobject{currentmarker}{\pgfqpoint{0.000000in}{-0.055556in}}{\pgfqpoint{0.000000in}{0.000000in}}{%
\pgfpathmoveto{\pgfqpoint{0.000000in}{0.000000in}}%
\pgfpathlineto{\pgfqpoint{0.000000in}{-0.055556in}}%
\pgfusepath{stroke,fill}%
}%
\begin{pgfscope}%
\pgfsys@transformshift{1.375417in}{3.713750in}%
\pgfsys@useobject{currentmarker}{}%
\end{pgfscope}%
\end{pgfscope}%
\begin{pgfscope}%
\pgftext[x=1.375417in,y=0.398750in,,top]{\sffamily\fontsize{12.000000}{14.400000}\selectfont M-F}%
\end{pgfscope}%
\begin{pgfscope}%
\pgfsetbuttcap%
\pgfsetroundjoin%
\definecolor{currentfill}{rgb}{0.000000,0.000000,0.000000}%
\pgfsetfillcolor{currentfill}%
\pgfsetlinewidth{0.501875pt}%
\definecolor{currentstroke}{rgb}{0.000000,0.000000,0.000000}%
\pgfsetstrokecolor{currentstroke}%
\pgfsetdash{}{0pt}%
\pgfsys@defobject{currentmarker}{\pgfqpoint{0.000000in}{0.000000in}}{\pgfqpoint{0.000000in}{0.055556in}}{%
\pgfpathmoveto{\pgfqpoint{0.000000in}{0.000000in}}%
\pgfpathlineto{\pgfqpoint{0.000000in}{0.055556in}}%
\pgfusepath{stroke,fill}%
}%
\begin{pgfscope}%
\pgfsys@transformshift{1.918657in}{0.454306in}%
\pgfsys@useobject{currentmarker}{}%
\end{pgfscope}%
\end{pgfscope}%
\begin{pgfscope}%
\pgfsetbuttcap%
\pgfsetroundjoin%
\definecolor{currentfill}{rgb}{0.000000,0.000000,0.000000}%
\pgfsetfillcolor{currentfill}%
\pgfsetlinewidth{0.501875pt}%
\definecolor{currentstroke}{rgb}{0.000000,0.000000,0.000000}%
\pgfsetstrokecolor{currentstroke}%
\pgfsetdash{}{0pt}%
\pgfsys@defobject{currentmarker}{\pgfqpoint{0.000000in}{-0.055556in}}{\pgfqpoint{0.000000in}{0.000000in}}{%
\pgfpathmoveto{\pgfqpoint{0.000000in}{0.000000in}}%
\pgfpathlineto{\pgfqpoint{0.000000in}{-0.055556in}}%
\pgfusepath{stroke,fill}%
}%
\begin{pgfscope}%
\pgfsys@transformshift{1.918657in}{3.713750in}%
\pgfsys@useobject{currentmarker}{}%
\end{pgfscope}%
\end{pgfscope}%
\begin{pgfscope}%
\pgftext[x=1.918657in,y=0.398750in,,top]{\sffamily\fontsize{12.000000}{14.400000}\selectfont P-W}%
\end{pgfscope}%
\begin{pgfscope}%
\pgfsetbuttcap%
\pgfsetroundjoin%
\definecolor{currentfill}{rgb}{0.000000,0.000000,0.000000}%
\pgfsetfillcolor{currentfill}%
\pgfsetlinewidth{0.501875pt}%
\definecolor{currentstroke}{rgb}{0.000000,0.000000,0.000000}%
\pgfsetstrokecolor{currentstroke}%
\pgfsetdash{}{0pt}%
\pgfsys@defobject{currentmarker}{\pgfqpoint{0.000000in}{0.000000in}}{\pgfqpoint{0.000000in}{0.055556in}}{%
\pgfpathmoveto{\pgfqpoint{0.000000in}{0.000000in}}%
\pgfpathlineto{\pgfqpoint{0.000000in}{0.055556in}}%
\pgfusepath{stroke,fill}%
}%
\begin{pgfscope}%
\pgfsys@transformshift{2.461898in}{0.454306in}%
\pgfsys@useobject{currentmarker}{}%
\end{pgfscope}%
\end{pgfscope}%
\begin{pgfscope}%
\pgfsetbuttcap%
\pgfsetroundjoin%
\definecolor{currentfill}{rgb}{0.000000,0.000000,0.000000}%
\pgfsetfillcolor{currentfill}%
\pgfsetlinewidth{0.501875pt}%
\definecolor{currentstroke}{rgb}{0.000000,0.000000,0.000000}%
\pgfsetstrokecolor{currentstroke}%
\pgfsetdash{}{0pt}%
\pgfsys@defobject{currentmarker}{\pgfqpoint{0.000000in}{-0.055556in}}{\pgfqpoint{0.000000in}{0.000000in}}{%
\pgfpathmoveto{\pgfqpoint{0.000000in}{0.000000in}}%
\pgfpathlineto{\pgfqpoint{0.000000in}{-0.055556in}}%
\pgfusepath{stroke,fill}%
}%
\begin{pgfscope}%
\pgfsys@transformshift{2.461898in}{3.713750in}%
\pgfsys@useobject{currentmarker}{}%
\end{pgfscope}%
\end{pgfscope}%
\begin{pgfscope}%
\pgftext[x=2.461898in,y=0.398750in,,top]{\sffamily\fontsize{12.000000}{14.400000}\selectfont C}%
\end{pgfscope}%
\begin{pgfscope}%
\pgfsetbuttcap%
\pgfsetroundjoin%
\definecolor{currentfill}{rgb}{0.000000,0.000000,0.000000}%
\pgfsetfillcolor{currentfill}%
\pgfsetlinewidth{0.501875pt}%
\definecolor{currentstroke}{rgb}{0.000000,0.000000,0.000000}%
\pgfsetstrokecolor{currentstroke}%
\pgfsetdash{}{0pt}%
\pgfsys@defobject{currentmarker}{\pgfqpoint{0.000000in}{0.000000in}}{\pgfqpoint{0.000000in}{0.055556in}}{%
\pgfpathmoveto{\pgfqpoint{0.000000in}{0.000000in}}%
\pgfpathlineto{\pgfqpoint{0.000000in}{0.055556in}}%
\pgfusepath{stroke,fill}%
}%
\begin{pgfscope}%
\pgfsys@transformshift{3.005139in}{0.454306in}%
\pgfsys@useobject{currentmarker}{}%
\end{pgfscope}%
\end{pgfscope}%
\begin{pgfscope}%
\pgfsetbuttcap%
\pgfsetroundjoin%
\definecolor{currentfill}{rgb}{0.000000,0.000000,0.000000}%
\pgfsetfillcolor{currentfill}%
\pgfsetlinewidth{0.501875pt}%
\definecolor{currentstroke}{rgb}{0.000000,0.000000,0.000000}%
\pgfsetstrokecolor{currentstroke}%
\pgfsetdash{}{0pt}%
\pgfsys@defobject{currentmarker}{\pgfqpoint{0.000000in}{-0.055556in}}{\pgfqpoint{0.000000in}{0.000000in}}{%
\pgfpathmoveto{\pgfqpoint{0.000000in}{0.000000in}}%
\pgfpathlineto{\pgfqpoint{0.000000in}{-0.055556in}}%
\pgfusepath{stroke,fill}%
}%
\begin{pgfscope}%
\pgfsys@transformshift{3.005139in}{3.713750in}%
\pgfsys@useobject{currentmarker}{}%
\end{pgfscope}%
\end{pgfscope}%
\begin{pgfscope}%
\pgftext[x=3.005139in,y=0.398750in,,top]{\sffamily\fontsize{12.000000}{14.400000}\selectfont R}%
\end{pgfscope}%
\begin{pgfscope}%
\pgfsetbuttcap%
\pgfsetroundjoin%
\definecolor{currentfill}{rgb}{0.000000,0.000000,0.000000}%
\pgfsetfillcolor{currentfill}%
\pgfsetlinewidth{0.501875pt}%
\definecolor{currentstroke}{rgb}{0.000000,0.000000,0.000000}%
\pgfsetstrokecolor{currentstroke}%
\pgfsetdash{}{0pt}%
\pgfsys@defobject{currentmarker}{\pgfqpoint{0.000000in}{0.000000in}}{\pgfqpoint{0.000000in}{0.055556in}}{%
\pgfpathmoveto{\pgfqpoint{0.000000in}{0.000000in}}%
\pgfpathlineto{\pgfqpoint{0.000000in}{0.055556in}}%
\pgfusepath{stroke,fill}%
}%
\begin{pgfscope}%
\pgfsys@transformshift{3.548380in}{0.454306in}%
\pgfsys@useobject{currentmarker}{}%
\end{pgfscope}%
\end{pgfscope}%
\begin{pgfscope}%
\pgfsetbuttcap%
\pgfsetroundjoin%
\definecolor{currentfill}{rgb}{0.000000,0.000000,0.000000}%
\pgfsetfillcolor{currentfill}%
\pgfsetlinewidth{0.501875pt}%
\definecolor{currentstroke}{rgb}{0.000000,0.000000,0.000000}%
\pgfsetstrokecolor{currentstroke}%
\pgfsetdash{}{0pt}%
\pgfsys@defobject{currentmarker}{\pgfqpoint{0.000000in}{-0.055556in}}{\pgfqpoint{0.000000in}{0.000000in}}{%
\pgfpathmoveto{\pgfqpoint{0.000000in}{0.000000in}}%
\pgfpathlineto{\pgfqpoint{0.000000in}{-0.055556in}}%
\pgfusepath{stroke,fill}%
}%
\begin{pgfscope}%
\pgfsys@transformshift{3.548380in}{3.713750in}%
\pgfsys@useobject{currentmarker}{}%
\end{pgfscope}%
\end{pgfscope}%
\begin{pgfscope}%
\pgftext[x=3.548380in,y=0.398750in,,top]{\sffamily\fontsize{12.000000}{14.400000}\selectfont T}%
\end{pgfscope}%
\begin{pgfscope}%
\pgftext[x=2.190278in,y=0.168661in,,top]{\sffamily\fontsize{12.000000}{14.400000}\selectfont Predicted label}%
\end{pgfscope}%
\begin{pgfscope}%
\pgfsetbuttcap%
\pgfsetroundjoin%
\definecolor{currentfill}{rgb}{0.000000,0.000000,0.000000}%
\pgfsetfillcolor{currentfill}%
\pgfsetlinewidth{0.501875pt}%
\definecolor{currentstroke}{rgb}{0.000000,0.000000,0.000000}%
\pgfsetstrokecolor{currentstroke}%
\pgfsetdash{}{0pt}%
\pgfsys@defobject{currentmarker}{\pgfqpoint{0.000000in}{0.000000in}}{\pgfqpoint{0.055556in}{0.000000in}}{%
\pgfpathmoveto{\pgfqpoint{0.000000in}{0.000000in}}%
\pgfpathlineto{\pgfqpoint{0.055556in}{0.000000in}}%
\pgfusepath{stroke,fill}%
}%
\begin{pgfscope}%
\pgfsys@transformshift{0.560556in}{3.442130in}%
\pgfsys@useobject{currentmarker}{}%
\end{pgfscope}%
\end{pgfscope}%
\begin{pgfscope}%
\pgfsetbuttcap%
\pgfsetroundjoin%
\definecolor{currentfill}{rgb}{0.000000,0.000000,0.000000}%
\pgfsetfillcolor{currentfill}%
\pgfsetlinewidth{0.501875pt}%
\definecolor{currentstroke}{rgb}{0.000000,0.000000,0.000000}%
\pgfsetstrokecolor{currentstroke}%
\pgfsetdash{}{0pt}%
\pgfsys@defobject{currentmarker}{\pgfqpoint{-0.055556in}{0.000000in}}{\pgfqpoint{0.000000in}{0.000000in}}{%
\pgfpathmoveto{\pgfqpoint{0.000000in}{0.000000in}}%
\pgfpathlineto{\pgfqpoint{-0.055556in}{0.000000in}}%
\pgfusepath{stroke,fill}%
}%
\begin{pgfscope}%
\pgfsys@transformshift{3.820000in}{3.442130in}%
\pgfsys@useobject{currentmarker}{}%
\end{pgfscope}%
\end{pgfscope}%
\begin{pgfscope}%
\pgftext[x=0.505000in,y=3.442130in,right,]{\sffamily\fontsize{12.000000}{14.400000}\selectfont U}%
\end{pgfscope}%
\begin{pgfscope}%
\pgfsetbuttcap%
\pgfsetroundjoin%
\definecolor{currentfill}{rgb}{0.000000,0.000000,0.000000}%
\pgfsetfillcolor{currentfill}%
\pgfsetlinewidth{0.501875pt}%
\definecolor{currentstroke}{rgb}{0.000000,0.000000,0.000000}%
\pgfsetstrokecolor{currentstroke}%
\pgfsetdash{}{0pt}%
\pgfsys@defobject{currentmarker}{\pgfqpoint{0.000000in}{0.000000in}}{\pgfqpoint{0.055556in}{0.000000in}}{%
\pgfpathmoveto{\pgfqpoint{0.000000in}{0.000000in}}%
\pgfpathlineto{\pgfqpoint{0.055556in}{0.000000in}}%
\pgfusepath{stroke,fill}%
}%
\begin{pgfscope}%
\pgfsys@transformshift{0.560556in}{2.898889in}%
\pgfsys@useobject{currentmarker}{}%
\end{pgfscope}%
\end{pgfscope}%
\begin{pgfscope}%
\pgfsetbuttcap%
\pgfsetroundjoin%
\definecolor{currentfill}{rgb}{0.000000,0.000000,0.000000}%
\pgfsetfillcolor{currentfill}%
\pgfsetlinewidth{0.501875pt}%
\definecolor{currentstroke}{rgb}{0.000000,0.000000,0.000000}%
\pgfsetstrokecolor{currentstroke}%
\pgfsetdash{}{0pt}%
\pgfsys@defobject{currentmarker}{\pgfqpoint{-0.055556in}{0.000000in}}{\pgfqpoint{0.000000in}{0.000000in}}{%
\pgfpathmoveto{\pgfqpoint{0.000000in}{0.000000in}}%
\pgfpathlineto{\pgfqpoint{-0.055556in}{0.000000in}}%
\pgfusepath{stroke,fill}%
}%
\begin{pgfscope}%
\pgfsys@transformshift{3.820000in}{2.898889in}%
\pgfsys@useobject{currentmarker}{}%
\end{pgfscope}%
\end{pgfscope}%
\begin{pgfscope}%
\pgftext[x=0.505000in,y=2.898889in,right,]{\sffamily\fontsize{12.000000}{14.400000}\selectfont M-F}%
\end{pgfscope}%
\begin{pgfscope}%
\pgfsetbuttcap%
\pgfsetroundjoin%
\definecolor{currentfill}{rgb}{0.000000,0.000000,0.000000}%
\pgfsetfillcolor{currentfill}%
\pgfsetlinewidth{0.501875pt}%
\definecolor{currentstroke}{rgb}{0.000000,0.000000,0.000000}%
\pgfsetstrokecolor{currentstroke}%
\pgfsetdash{}{0pt}%
\pgfsys@defobject{currentmarker}{\pgfqpoint{0.000000in}{0.000000in}}{\pgfqpoint{0.055556in}{0.000000in}}{%
\pgfpathmoveto{\pgfqpoint{0.000000in}{0.000000in}}%
\pgfpathlineto{\pgfqpoint{0.055556in}{0.000000in}}%
\pgfusepath{stroke,fill}%
}%
\begin{pgfscope}%
\pgfsys@transformshift{0.560556in}{2.355648in}%
\pgfsys@useobject{currentmarker}{}%
\end{pgfscope}%
\end{pgfscope}%
\begin{pgfscope}%
\pgfsetbuttcap%
\pgfsetroundjoin%
\definecolor{currentfill}{rgb}{0.000000,0.000000,0.000000}%
\pgfsetfillcolor{currentfill}%
\pgfsetlinewidth{0.501875pt}%
\definecolor{currentstroke}{rgb}{0.000000,0.000000,0.000000}%
\pgfsetstrokecolor{currentstroke}%
\pgfsetdash{}{0pt}%
\pgfsys@defobject{currentmarker}{\pgfqpoint{-0.055556in}{0.000000in}}{\pgfqpoint{0.000000in}{0.000000in}}{%
\pgfpathmoveto{\pgfqpoint{0.000000in}{0.000000in}}%
\pgfpathlineto{\pgfqpoint{-0.055556in}{0.000000in}}%
\pgfusepath{stroke,fill}%
}%
\begin{pgfscope}%
\pgfsys@transformshift{3.820000in}{2.355648in}%
\pgfsys@useobject{currentmarker}{}%
\end{pgfscope}%
\end{pgfscope}%
\begin{pgfscope}%
\pgftext[x=0.505000in,y=2.355648in,right,]{\sffamily\fontsize{12.000000}{14.400000}\selectfont P-W}%
\end{pgfscope}%
\begin{pgfscope}%
\pgfsetbuttcap%
\pgfsetroundjoin%
\definecolor{currentfill}{rgb}{0.000000,0.000000,0.000000}%
\pgfsetfillcolor{currentfill}%
\pgfsetlinewidth{0.501875pt}%
\definecolor{currentstroke}{rgb}{0.000000,0.000000,0.000000}%
\pgfsetstrokecolor{currentstroke}%
\pgfsetdash{}{0pt}%
\pgfsys@defobject{currentmarker}{\pgfqpoint{0.000000in}{0.000000in}}{\pgfqpoint{0.055556in}{0.000000in}}{%
\pgfpathmoveto{\pgfqpoint{0.000000in}{0.000000in}}%
\pgfpathlineto{\pgfqpoint{0.055556in}{0.000000in}}%
\pgfusepath{stroke,fill}%
}%
\begin{pgfscope}%
\pgfsys@transformshift{0.560556in}{1.812407in}%
\pgfsys@useobject{currentmarker}{}%
\end{pgfscope}%
\end{pgfscope}%
\begin{pgfscope}%
\pgfsetbuttcap%
\pgfsetroundjoin%
\definecolor{currentfill}{rgb}{0.000000,0.000000,0.000000}%
\pgfsetfillcolor{currentfill}%
\pgfsetlinewidth{0.501875pt}%
\definecolor{currentstroke}{rgb}{0.000000,0.000000,0.000000}%
\pgfsetstrokecolor{currentstroke}%
\pgfsetdash{}{0pt}%
\pgfsys@defobject{currentmarker}{\pgfqpoint{-0.055556in}{0.000000in}}{\pgfqpoint{0.000000in}{0.000000in}}{%
\pgfpathmoveto{\pgfqpoint{0.000000in}{0.000000in}}%
\pgfpathlineto{\pgfqpoint{-0.055556in}{0.000000in}}%
\pgfusepath{stroke,fill}%
}%
\begin{pgfscope}%
\pgfsys@transformshift{3.820000in}{1.812407in}%
\pgfsys@useobject{currentmarker}{}%
\end{pgfscope}%
\end{pgfscope}%
\begin{pgfscope}%
\pgftext[x=0.505000in,y=1.812407in,right,]{\sffamily\fontsize{12.000000}{14.400000}\selectfont C}%
\end{pgfscope}%
\begin{pgfscope}%
\pgfsetbuttcap%
\pgfsetroundjoin%
\definecolor{currentfill}{rgb}{0.000000,0.000000,0.000000}%
\pgfsetfillcolor{currentfill}%
\pgfsetlinewidth{0.501875pt}%
\definecolor{currentstroke}{rgb}{0.000000,0.000000,0.000000}%
\pgfsetstrokecolor{currentstroke}%
\pgfsetdash{}{0pt}%
\pgfsys@defobject{currentmarker}{\pgfqpoint{0.000000in}{0.000000in}}{\pgfqpoint{0.055556in}{0.000000in}}{%
\pgfpathmoveto{\pgfqpoint{0.000000in}{0.000000in}}%
\pgfpathlineto{\pgfqpoint{0.055556in}{0.000000in}}%
\pgfusepath{stroke,fill}%
}%
\begin{pgfscope}%
\pgfsys@transformshift{0.560556in}{1.269167in}%
\pgfsys@useobject{currentmarker}{}%
\end{pgfscope}%
\end{pgfscope}%
\begin{pgfscope}%
\pgfsetbuttcap%
\pgfsetroundjoin%
\definecolor{currentfill}{rgb}{0.000000,0.000000,0.000000}%
\pgfsetfillcolor{currentfill}%
\pgfsetlinewidth{0.501875pt}%
\definecolor{currentstroke}{rgb}{0.000000,0.000000,0.000000}%
\pgfsetstrokecolor{currentstroke}%
\pgfsetdash{}{0pt}%
\pgfsys@defobject{currentmarker}{\pgfqpoint{-0.055556in}{0.000000in}}{\pgfqpoint{0.000000in}{0.000000in}}{%
\pgfpathmoveto{\pgfqpoint{0.000000in}{0.000000in}}%
\pgfpathlineto{\pgfqpoint{-0.055556in}{0.000000in}}%
\pgfusepath{stroke,fill}%
}%
\begin{pgfscope}%
\pgfsys@transformshift{3.820000in}{1.269167in}%
\pgfsys@useobject{currentmarker}{}%
\end{pgfscope}%
\end{pgfscope}%
\begin{pgfscope}%
\pgftext[x=0.505000in,y=1.269167in,right,]{\sffamily\fontsize{12.000000}{14.400000}\selectfont R}%
\end{pgfscope}%
\begin{pgfscope}%
\pgfsetbuttcap%
\pgfsetroundjoin%
\definecolor{currentfill}{rgb}{0.000000,0.000000,0.000000}%
\pgfsetfillcolor{currentfill}%
\pgfsetlinewidth{0.501875pt}%
\definecolor{currentstroke}{rgb}{0.000000,0.000000,0.000000}%
\pgfsetstrokecolor{currentstroke}%
\pgfsetdash{}{0pt}%
\pgfsys@defobject{currentmarker}{\pgfqpoint{0.000000in}{0.000000in}}{\pgfqpoint{0.055556in}{0.000000in}}{%
\pgfpathmoveto{\pgfqpoint{0.000000in}{0.000000in}}%
\pgfpathlineto{\pgfqpoint{0.055556in}{0.000000in}}%
\pgfusepath{stroke,fill}%
}%
\begin{pgfscope}%
\pgfsys@transformshift{0.560556in}{0.725926in}%
\pgfsys@useobject{currentmarker}{}%
\end{pgfscope}%
\end{pgfscope}%
\begin{pgfscope}%
\pgfsetbuttcap%
\pgfsetroundjoin%
\definecolor{currentfill}{rgb}{0.000000,0.000000,0.000000}%
\pgfsetfillcolor{currentfill}%
\pgfsetlinewidth{0.501875pt}%
\definecolor{currentstroke}{rgb}{0.000000,0.000000,0.000000}%
\pgfsetstrokecolor{currentstroke}%
\pgfsetdash{}{0pt}%
\pgfsys@defobject{currentmarker}{\pgfqpoint{-0.055556in}{0.000000in}}{\pgfqpoint{0.000000in}{0.000000in}}{%
\pgfpathmoveto{\pgfqpoint{0.000000in}{0.000000in}}%
\pgfpathlineto{\pgfqpoint{-0.055556in}{0.000000in}}%
\pgfusepath{stroke,fill}%
}%
\begin{pgfscope}%
\pgfsys@transformshift{3.820000in}{0.725926in}%
\pgfsys@useobject{currentmarker}{}%
\end{pgfscope}%
\end{pgfscope}%
\begin{pgfscope}%
\pgftext[x=0.505000in,y=0.725926in,right,]{\sffamily\fontsize{12.000000}{14.400000}\selectfont T}%
\end{pgfscope}%
\begin{pgfscope}%
\pgftext[x=0.124419in,y=2.084028in,,bottom,rotate=90.000000]{\sffamily\fontsize{12.000000}{14.400000}\selectfont True label}%
\end{pgfscope}%
\begin{pgfscope}%
\definecolor{textcolor}{rgb}{1.000000,1.000000,1.000000}%
\pgfsetstrokecolor{textcolor}%
\pgfsetfillcolor{textcolor}%
\pgftext[x=0.832176in,y=3.442130in,,base]{\color{textcolor}\sffamily\fontsize{12.000000}{14.400000}\selectfont 143}%
\end{pgfscope}%
\begin{pgfscope}%
\pgftext[x=1.375417in,y=3.442130in,,base]{\sffamily\fontsize{12.000000}{14.400000}\selectfont 15}%
\end{pgfscope}%
\begin{pgfscope}%
\pgftext[x=1.918657in,y=3.442130in,,base]{\sffamily\fontsize{12.000000}{14.400000}\selectfont 9}%
\end{pgfscope}%
\begin{pgfscope}%
\pgftext[x=2.461898in,y=3.442130in,,base]{\sffamily\fontsize{12.000000}{14.400000}\selectfont 6}%
\end{pgfscope}%
\begin{pgfscope}%
\pgftext[x=3.005139in,y=3.442130in,,base]{\sffamily\fontsize{12.000000}{14.400000}\selectfont 1}%
\end{pgfscope}%
\begin{pgfscope}%
\pgftext[x=3.548380in,y=3.442130in,,base]{\sffamily\fontsize{12.000000}{14.400000}\selectfont 1}%
\end{pgfscope}%
\begin{pgfscope}%
\pgftext[x=0.832176in,y=2.898889in,,base]{\sffamily\fontsize{12.000000}{14.400000}\selectfont 13}%
\end{pgfscope}%
\begin{pgfscope}%
\pgftext[x=1.375417in,y=2.898889in,,base]{\sffamily\fontsize{12.000000}{14.400000}\selectfont 38}%
\end{pgfscope}%
\begin{pgfscope}%
\pgftext[x=1.918657in,y=2.898889in,,base]{\sffamily\fontsize{12.000000}{14.400000}\selectfont 10}%
\end{pgfscope}%
\begin{pgfscope}%
\pgftext[x=2.461898in,y=2.898889in,,base]{\sffamily\fontsize{12.000000}{14.400000}\selectfont 0}%
\end{pgfscope}%
\begin{pgfscope}%
\pgftext[x=3.005139in,y=2.898889in,,base]{\sffamily\fontsize{12.000000}{14.400000}\selectfont 3}%
\end{pgfscope}%
\begin{pgfscope}%
\pgftext[x=3.548380in,y=2.898889in,,base]{\sffamily\fontsize{12.000000}{14.400000}\selectfont 1}%
\end{pgfscope}%
\begin{pgfscope}%
\pgftext[x=0.832176in,y=2.355648in,,base]{\sffamily\fontsize{12.000000}{14.400000}\selectfont 15}%
\end{pgfscope}%
\begin{pgfscope}%
\pgftext[x=1.375417in,y=2.355648in,,base]{\sffamily\fontsize{12.000000}{14.400000}\selectfont 8}%
\end{pgfscope}%
\begin{pgfscope}%
\pgftext[x=1.918657in,y=2.355648in,,base]{\sffamily\fontsize{12.000000}{14.400000}\selectfont 44}%
\end{pgfscope}%
\begin{pgfscope}%
\pgftext[x=2.461898in,y=2.355648in,,base]{\sffamily\fontsize{12.000000}{14.400000}\selectfont 3}%
\end{pgfscope}%
\begin{pgfscope}%
\pgftext[x=3.005139in,y=2.355648in,,base]{\sffamily\fontsize{12.000000}{14.400000}\selectfont 0}%
\end{pgfscope}%
\begin{pgfscope}%
\pgftext[x=3.548380in,y=2.355648in,,base]{\sffamily\fontsize{12.000000}{14.400000}\selectfont 0}%
\end{pgfscope}%
\begin{pgfscope}%
\pgftext[x=0.832176in,y=1.812407in,,base]{\sffamily\fontsize{12.000000}{14.400000}\selectfont 1}%
\end{pgfscope}%
\begin{pgfscope}%
\pgftext[x=1.375417in,y=1.812407in,,base]{\sffamily\fontsize{12.000000}{14.400000}\selectfont 1}%
\end{pgfscope}%
\begin{pgfscope}%
\pgftext[x=1.918657in,y=1.812407in,,base]{\sffamily\fontsize{12.000000}{14.400000}\selectfont 2}%
\end{pgfscope}%
\begin{pgfscope}%
\pgftext[x=2.461898in,y=1.812407in,,base]{\sffamily\fontsize{12.000000}{14.400000}\selectfont 14}%
\end{pgfscope}%
\begin{pgfscope}%
\pgftext[x=3.005139in,y=1.812407in,,base]{\sffamily\fontsize{12.000000}{14.400000}\selectfont 3}%
\end{pgfscope}%
\begin{pgfscope}%
\pgftext[x=3.548380in,y=1.812407in,,base]{\sffamily\fontsize{12.000000}{14.400000}\selectfont 0}%
\end{pgfscope}%
\begin{pgfscope}%
\pgftext[x=0.832176in,y=1.269167in,,base]{\sffamily\fontsize{12.000000}{14.400000}\selectfont 4}%
\end{pgfscope}%
\begin{pgfscope}%
\pgftext[x=1.375417in,y=1.269167in,,base]{\sffamily\fontsize{12.000000}{14.400000}\selectfont 2}%
\end{pgfscope}%
\begin{pgfscope}%
\pgftext[x=1.918657in,y=1.269167in,,base]{\sffamily\fontsize{12.000000}{14.400000}\selectfont 0}%
\end{pgfscope}%
\begin{pgfscope}%
\pgftext[x=2.461898in,y=1.269167in,,base]{\sffamily\fontsize{12.000000}{14.400000}\selectfont 1}%
\end{pgfscope}%
\begin{pgfscope}%
\pgftext[x=3.005139in,y=1.269167in,,base]{\sffamily\fontsize{12.000000}{14.400000}\selectfont 13}%
\end{pgfscope}%
\begin{pgfscope}%
\pgftext[x=3.548380in,y=1.269167in,,base]{\sffamily\fontsize{12.000000}{14.400000}\selectfont 0}%
\end{pgfscope}%
\begin{pgfscope}%
\pgftext[x=0.832176in,y=0.725926in,,base]{\sffamily\fontsize{12.000000}{14.400000}\selectfont 2}%
\end{pgfscope}%
\begin{pgfscope}%
\pgftext[x=1.375417in,y=0.725926in,,base]{\sffamily\fontsize{12.000000}{14.400000}\selectfont 0}%
\end{pgfscope}%
\begin{pgfscope}%
\pgftext[x=1.918657in,y=0.725926in,,base]{\sffamily\fontsize{12.000000}{14.400000}\selectfont 0}%
\end{pgfscope}%
\begin{pgfscope}%
\pgftext[x=2.461898in,y=0.725926in,,base]{\sffamily\fontsize{12.000000}{14.400000}\selectfont 0}%
\end{pgfscope}%
\begin{pgfscope}%
\pgftext[x=3.005139in,y=0.725926in,,base]{\sffamily\fontsize{12.000000}{14.400000}\selectfont 0}%
\end{pgfscope}%
\begin{pgfscope}%
\pgftext[x=3.548380in,y=0.725926in,,base]{\sffamily\fontsize{12.000000}{14.400000}\selectfont 1}%
\end{pgfscope}%
\end{pgfpicture}%
\makeatother%
\endgroup%

%% file: semeval2018.bbl
\begin{thebibliography}{17}
\expandafter\ifx\csname natexlab\endcsname\relax\def\natexlab#1{#1}\fi

\bibitem[{Ammar et~al.(2017)Ammar, Peters, Bhagavatula, and
  Power}]{Ammar2017TheAS}
Waleed Ammar, Matthew~E. Peters, Chandra Bhagavatula, and Russell Power. 2017.
\newblock The {AI2} system at {SemEval-2017} {T}ask 10 ({ScienceIE}):
  semi-supervised end-to-end entity and relation extraction.
\newblock In \emph{SemEval-2017}.

\bibitem[{Augenstein et~al.(2017)Augenstein, Das, Riedel, Vikraman, and
  McCallum}]{Augenstein2017SemEval2T}
Isabelle Augenstein, Mrinal Das, Sebastian Riedel, Lakshmi Vikraman, and
  Andrew~D McCallum. 2017.
\newblock {SemEval} 2017 {T}ask 10: {ScienceIE} - extracting keyphrases and
  relations from scientific publications.
\newblock In \emph{SemEval-2017}.

\bibitem[{Bird et~al.(2008)Bird, Dale, Dorr, Gibson, Joseph, Kan, Lee, Powley,
  Radev, and Tan}]{Bird2008TheAA}
Steven Bird, Robert Dale, Bonnie~J. Dorr, Bryan~R. Gibson, Mark~Thomas Joseph,
  Min-Yen Kan, Dongwon Lee, Brett Powley, Dragomir~R. Radev, and Yee~Fan Tan.
  2008.
\newblock The {ACL} anthology reference corpus: A reference dataset for
  bibliographic research in computational linguistics.
\newblock In \emph{LREC}.

\bibitem[{Cohan et~al.(2018)Cohan, Dernoncourt, Kim, Bui, Chang, and
  Goharian}]{cohan2018}
Arman Cohan, Franck Dernoncourt, Doo~Soon Kim, Kim~Seokhwan Bui, Trung, Walter
  Chang, and Nazli Goharian. 2018.
\newblock A discourse-aware attention model for abstractive summarization of
  long documents.
\newblock In \emph{NAACL-HLT}.

\bibitem[{G\'{a}bor et~al.(2018)G\'{a}bor, Buscaldi, Schumann, QasemiZadeh,
  Zargayouna, and Charnois}]{SemEval2018Task7}
Kata G\'{a}bor, Davide Buscaldi, Anne-Kathrin Schumann, Behrang QasemiZadeh,
  Ha\"{i}fa Zargayouna, and Thierry Charnois. 2018.
\newblock {SemEval-2018} {T}ask 7: Semantic relation extraction and
  classification in scientific papers.
\newblock In \emph{SemEval-2018}.

\bibitem[{G{\'a}bor et~al.(2016)G{\'a}bor, Zargayouna, Buscaldi, Tellier, and
  Charnois}]{Gbor2016SemanticAO}
Kata G{\'a}bor, Ha{\"i}fa Zargayouna, Davide Buscaldi, Isabelle Tellier, and
  Thierry Charnois. 2016.
\newblock Semantic annotation of the {ACL} anthology corpus for the automatic
  analysis of scientific literature.
\newblock In \emph{LREC}.

\bibitem[{Hochreiter and Schmidhuber(1997)}]{hochreiter1997long}
Sepp Hochreiter and J{\"u}rgen Schmidhuber. 1997.
\newblock Long short-term memory.
\newblock \emph{Neural computation}, 9(8):1735--1780.

\bibitem[{Lee et~al.(2017{\natexlab{a}})Lee, Dernoncourt, and
  Szolovits}]{lee-dernoncourt-szolovits:2017:SemEval}
Ji~Young Lee, Franck Dernoncourt, and Peter Szolovits. 2017{\natexlab{a}}.
\newblock {MIT} at {SemEval-2017} {T}ask 10: Relation extraction with
  convolutional neural networks.
\newblock In \emph{SemEval-2017}.

\bibitem[{Lee et~al.(2017{\natexlab{b}})Lee, Lee, and Tseng}]{Lee2017TheNS}
Lung-Hao Lee, Kuei-Ching Lee, and Y~Jane Tseng. 2017{\natexlab{b}}.
\newblock The {NTNU} system at {SemEval-2017} {T}ask 10: Extracting keyphrases
  and relations from scientific publications using multiple conditional random
  fields.
\newblock In \emph{SemEval-2017}.

\bibitem[{Luan et~al.(2017)Luan, Ostendorf, and
  Hajishirzi}]{Luan2017ScientificIE}
Yi~Luan, Mari Ostendorf, and Hannaneh Hajishirzi. 2017.
\newblock Scientific information extraction with semi-supervised neural
  tagging.
\newblock In \emph{EMNLP}.

\bibitem[{MacAvaney et~al.(2017)MacAvaney, Cohan, and
  Goharian}]{macavaney2017guir}
Sean MacAvaney, Arman Cohan, and Nazli Goharian. 2017.
\newblock {GUIR} at {SemEval-2017} {T}ask 12: A framework for cross-domain
  clinical temporal information extraction.
\newblock In \emph{SemEval-2017}.

\bibitem[{Mikolov et~al.(2018)Mikolov, Grave, Bojanowski, Puhrsch, and
  Joulin}]{mikolov2018advances}
Tomas Mikolov, Edouard Grave, Piotr Bojanowski, Christian Puhrsch, and Armand
  Joulin. 2018.
\newblock Advances in pre-training distributed word representations.
\newblock In \emph{LREC}.

\bibitem[{Mintz et~al.(2009)Mintz, Bills, Snow, and
  Jurafsky}]{mintz2009distant}
Mike Mintz, Steven Bills, Rion Snow, and Dan Jurafsky. 2009.
\newblock Distant supervision for relation extraction without labeled data.
\newblock In \emph{ACL/IJCNLP}. Association for Computational Linguistics.

\bibitem[{Miwa and Bansal(2016)}]{miwa2016end}
Makoto Miwa and Mohit Bansal. 2016.
\newblock End-to-end relation extraction using lstms on sequences and tree
  structures.
\newblock In \emph{ACL}.

\bibitem[{Tai et~al.(2015)Tai, Socher, and Manning}]{tai2015improved}
Kai~Sheng Tai, Richard Socher, and Christopher~D Manning. 2015.
\newblock Improved semantic representations from tree-structured long
  short-term memory networks.
\newblock \emph{arXiv preprint arXiv:1503.00075}.

\bibitem[{Xu et~al.(2015)Xu, Mou, Li, Chen, Peng, and Jin}]{xu2015classifying}
Yan Xu, Lili Mou, Ge~Li, Yunchuan Chen, Hao Peng, and Zhi Jin. 2015.
\newblock Classifying relations via long short term memory networks along
  shortest dependency paths.
\newblock In \emph{EMNLP}.

\bibitem[{Zheng et~al.(2014)Zheng, Howsmon, Zhang, Hahn, McGuinness, Hendler,
  and Ji}]{Zheng2014EntityLF}
Jinguang Zheng, Daniel~P Howsmon, Boliang Zhang, Juergen Hahn, Deborah~L.
  McGuinness, James~A. Hendler, and Heng Ji. 2014.
\newblock Entity linking for biomedical literature.
\newblock In \emph{DTMBIO@CIKM}.

\end{thebibliography}
